\documentclass[11pt,a4paper]{article}
\usepackage[hyperref]{emnlp-ijcnlp-2019}
\usepackage{times}
\usepackage{latexsym}

\usepackage{url}
\usepackage{amssymb}
\usepackage{amsmath}
\usepackage{graphicx}

\aclfinalcopy 

\title{Neural Cross-Lingual Relation Extraction Based on \\ Bilingual Word Embedding Mapping}

\author{Jian Ni \and Radu Florian\\
    IBM Research AI\\
    1101 Kitchawan Road, Yorktown Heights, NY 10598, USA\\
  {\tt \{nij, raduf\}@us.ibm.com}}

\date{}

\begin{document}

\maketitle

\begin{abstract}
Relation extraction (RE) seeks to detect and classify semantic relationships between entities, which provides useful information for many NLP applications. 
Since the state-of-the-art RE models require large amounts of manually annotated data and language-specific resources to achieve high accuracy, it is very challenging to transfer an RE model of a resource-rich language to a resource-poor language. In this paper, we propose a new approach for cross-lingual RE model transfer based on bilingual word embedding mapping. It projects word embeddings from a target language to a source language, so that a well-trained source-language neural network RE model can be directly applied to the target language. Experiment results show that the proposed approach achieves very good performance for a number of target languages on both in-house and open datasets, using a small bilingual dictionary with only 1K word pairs.
\end{abstract}

\section{Introduction}

\emph{Relation extraction} (RE) is an important information extraction task that seeks to detect and classify semantic relationships 
between entities like persons, organizations, geo-political entities, locations, and events. It provides useful information for many NLP applications such as knowledge base construction, text mining and question answering. For example, the entity \emph{Washington, D.C.} and the entity \emph{United States} have a \emph{CapitalOf} relationship, and extraction of such relationships can help answer questions like ``What is the capital city of the United States?"

Traditional RE models (e.g., \citet{Zelenko2003,Kambhatla2004,Li2014}) require careful feature engineering to derive and combine various lexical, syntactic and semantic features. Recently, neural network RE models (e.g., \citet{Zeng2014,Santos2015,Miwa2016,Nguyen2016}) have become very successful. These models employ a certain level of automatic feature learning by using
word embeddings, which significantly simplifies the feature engineering task while considerably improving the accuracy, achieving the state-of-the-art performance for relation extraction.

All the above RE models are supervised machine learning models that need to be trained with large amounts of manually annotated RE data to achieve high accuracy. However, annotating RE data by human is expensive and time-consuming, and can be quite difficult for a new
language. Moreover, most RE models require language-specific resources such as dependency parsers and part-of-speech (POS) taggers,
which also makes it very challenging to transfer an RE model of a resource-rich language to a resource-poor language.

There are a few existing weakly supervised cross-lingual RE approaches that require no human annotation in the target languages, e.g.,
\citet{Kim2010,Kim2012,Faruqui2015,Zou2018}. However, the existing approaches require aligned parallel corpora or machine translation systems,
which may not be readily available in practice.

In this paper, we make the following contributions to cross-lingual RE:
\begin{itemize}

\item We propose a new approach for direct cross-lingual RE model transfer based on bilingual word embedding mapping. It projects word embeddings from a target language to a source language (e.g., English), so that a well-trained
source-language RE model can be directly applied to the target language, with no manually annotated RE data needed for the target language. 

\item We design a deep neural network architecture for the source-language (English) RE model that uses word embeddings and generic
language-independent features as the input. The English RE model achieves the-state-of-the-art performance without using
language-specific resources.

\item We conduct extensive experiments which show that the proposed approach achieves very good performance (up to $79\%$ of the accuracy of the supervised target-language RE model) for a number of target languages on both in-house and the ACE05 datasets \cite{Walker2006},
using a small bilingual dictionary with only 1K word pairs. To the best of our knowledge, this is the first work that includes empirical studies for cross-lingual RE on several languages across a variety of language families, without using aligned parallel corpora or machine translation systems.

\end{itemize}

We organize the paper as follows. In Section 2 we provide an overview of our approach. In Section 3 we describe how to build monolingual
word embeddings and learn a linear mapping between two languages. In Section 4 we present a neural network architecture for the
source-language (English). In Section 5 we evaluate the performance of the proposed approach for a number of target languages. We
discuss related work in Section 6 and conclude the paper in Section 7.

\begin{figure*}
\includegraphics[scale=0.5]{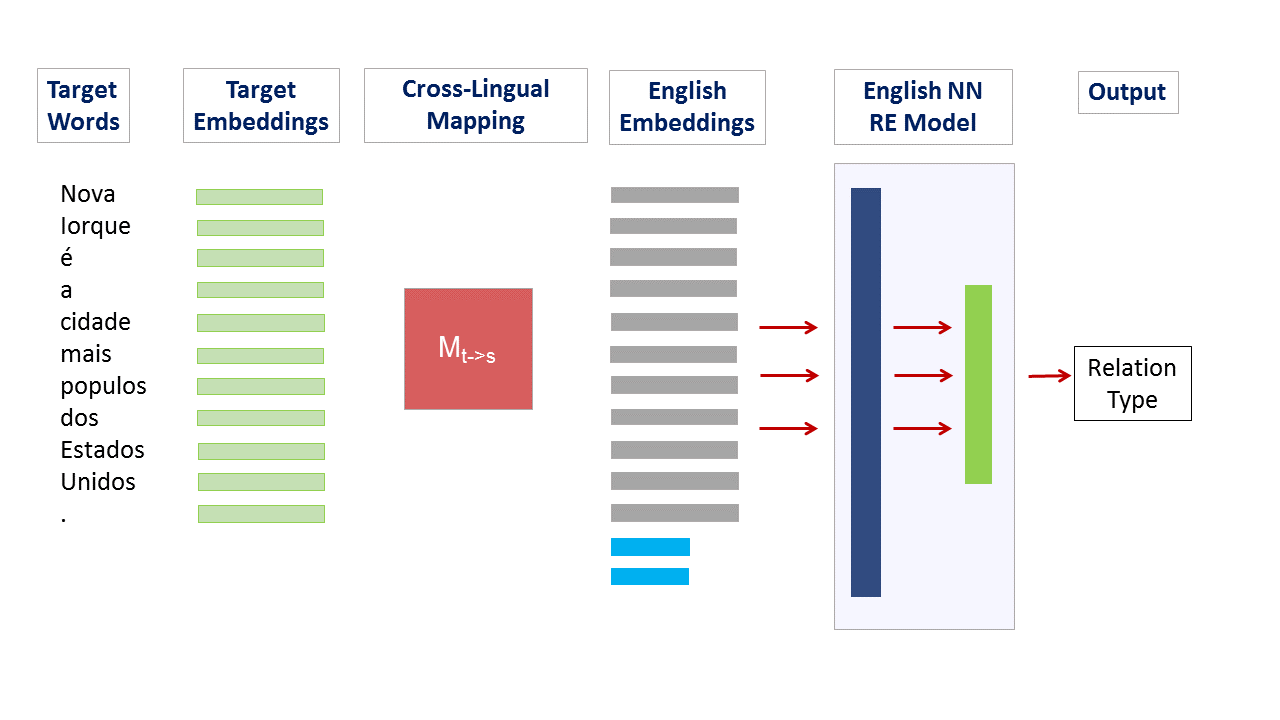} \caption{Neural cross-lingual relation extraction based on bilingual word embedding mapping - target language: Portuguese, source language: English.} \label{fig:cross-lingual-RE}
\end{figure*}

\section{Overview of the Approach}

We summarize the main steps of our neural cross-lingual RE model transfer approach as follows.

\begin{itemize}
\item[1.] Build word embeddings for the source language and the target language separately using monolingual data.

\item[2.] Learn a linear mapping that projects the target-language word embeddings into the source-language embedding space using
a small bilingual dictionary.

\item[3.] Build a neural network source-language RE model that uses word embeddings and generic language-independent features as the
input.

\item[4.] For a target-language sentence and any two entities in it, project the word embeddings of the words in the sentence to the
source-language word embeddings using the linear mapping, and then apply the source-language RE model on the projected word embeddings
to classify the relationship between the two entities. An example is shown in Figure \ref{fig:cross-lingual-RE}, where the target
language is Portuguese and the source language is English.
\end{itemize}

 We will describe each component of our approach in the subsequent sections.

\section{Cross-Lingual Word Embeddings} \label{section:cross-lingual-embeddings}

In recent years, vector representations of words, known as \emph{word embeddings}, become ubiquitous for many NLP applications
\cite{Collobert2011,Mikolov2013a,Pennington2014}.

A monolingual word embedding model maps words in the vocabulary $\mathcal{V}$ of a language to real-valued vectors in
$\mathbb{R}^{d\times 1}$. The dimension of the vector space $d$ is normally much smaller than the size of the vocabulary
$V=|\mathcal{V}|$ for efficient representation. It also aims to capture semantic similarities between the words based on their
distributional properties in large samples of monolingual data.

Cross-lingual word embedding models try to build word embeddings across multiple languages \cite{Upadhyay2016,Ruder2017}. One approach builds monolingual word embeddings separately and then maps them to the same vector space using a bilingual dictionary \cite{Mikolov2013b,Faruqui2014}. Another approach builds multilingual word embeddings in a shared vector space simultaneously, by
generating mixed language corpora using aligned sentences \cite{Luong2015,Gouws2015a}.

In this paper, we adopt the technique in \cite{Mikolov2013b} because it only requires a small bilingual dictionary of aligned word
pairs, and does not require parallel corpora of aligned sentences which could be more difficult to obtain.

\subsection{Monolingual Word Embeddings} \label{subsection:monolingual_embeddings}

To build monolingual word embeddings for the source and target languages, we use a variant of the Continuous Bag-of-Words (CBOW)
word2vec model \cite{Mikolov2013a}. 

The standard CBOW model has two matrices, the input word matrix $\tilde{\mathbf{X}} \in
\mathbb{R}^{d\times V}$ and the output word matrix $\mathbf{X} \in \mathbb{R}^{d\times V}$. For the $i$th word $w_i$ in $\mathcal{V}$,
let $\mathbf{e}(w_i) \in \mathbb{R}^{V \times 1}$ be a one-hot vector with 1 at index $i$ and 0s at other indexes, so that $\tilde{\mathbf{x}}_i
= \tilde{\mathbf{X}}\mathbf{e}(w_i)$ (the $i$th column of $\tilde{\mathbf{X}}$) is the input vector representation of word $w_i$,
and $\mathbf{x}_i = \mathbf{X}\mathbf{e}(w_i)$ (the $i$th column of $\mathbf{X}$) is the output vector representation (i.e., word embedding) of word $w_i$.

Given a sequence of training words $w_1, w_2, ..., w_N$, the CBOW model seeks to predict a target word $w_t$ using a window of $2c$ context words surrounding $w_t$, by maximizing the following objective function:
\begin{eqnarray*}
\mathcal{L} = \frac{1}{N} \sum_{t=1}^{N} \log P(w_t | w_{t-c},...,w_{t-1},w_{t+1},...,w_{t+c}) \nonumber \\
\end{eqnarray*}
The conditional probability is calculated using a softmax function:
\begin{eqnarray} \label{eq:conditionalprob}
P(w_t | w_{t-c},...,w_{t+c}) = \frac{\exp(\mathbf{x}_t^\mathrm{T} \tilde{\mathbf{x}}_{c(t)})}{\sum_{i=1}^{V}\exp(\mathbf{x}_i^\mathrm{T}
\tilde{\mathbf{x}}_{c(t)})}
\end{eqnarray}
where $\mathbf{x}_t=\mathbf{X}\mathbf{e}(w_t)$ is the output vector representation of word $w_t$, and
\begin{eqnarray}
\tilde{\mathbf{x}}_{c(t)} = \sum_{-c \leq j \leq c, j\neq 0} \tilde{\mathbf{X}} \mathbf{e}(w_{t+j})
\end{eqnarray}
is the sum of the input vector representations of the context words.

In our variant of the CBOW model, we use a separate input word matrix $\tilde{\mathbf{X}}_j$ for a context word at position $j, -c \leq j \leq c,
j\neq 0$.  In addition, we employ weights that decay with the distances of the context words to the 
target word. Under these modifications, we have
\begin{equation}
 \tilde{\mathbf{x}}_{c(t)}^{new} = \sum_{-c \leq j
\leq c, j\neq 0} \frac{1}{|j|} \tilde{\mathbf{X}}_j \mathbf{e}(w_{t+j})
\end{equation}

We use the variant to build monolingual word embeddings because experiments on named entity recognition and word similarity tasks showed this variant leads to small improvements over the standard CBOW model \cite{Ni2017}.

\subsection{Bilingual Word Embedding Mapping} \label{subsection:cross-lingual-embedding-mapping}

\citet{Mikolov2013b} observed that word embeddings of different languages often have similar geometric arrangements, and suggested to
learn a linear mapping between the vector spaces.

Let $\mathcal{D}$ be a bilingual dictionary with aligned word pairs ($w_i, v_i)_{i=1,...,D}$ between a source language $s$ and a target
language $t$, where $w_i$ is a source-language word and $v_i$ is the translation of $w_i$ in the target language.  Let $\mathbf{x}_i \in
\mathbb{R}^{d \times 1}$ be the word embedding of the source-language word $w_i$, $\mathbf{y}_i \in \mathbb{R}^{d \times 1}$
be the word embedding of the target-language word $v_i$.

We find a linear mapping (matrix) $\mathbf{M}_{t\rightarrow
s}$ such that $\mathbf{M}_{t\rightarrow s}\mathbf{y}_i$ approximates $\mathbf{x}_i$, by solving the following least squares problem using the dictionary as the training set:
\begin{equation} \label{eq:bilingual-mapping}
\mathbf{M}_{t\rightarrow s} = \arg\min_{\mathbf{M} \in \mathbb{R}^{d \times d}} \sum_{i=1}^{D}|| \mathbf{x}_i -
\mathbf{M}\mathbf{y}_i||^2
\end{equation}

Using $\mathbf{M}_{t\rightarrow s}$, for any target-language word $v$ with word embedding $\mathbf{y}$, we can project it into
the source-language embedding space as $\mathbf{M}_{t\rightarrow s}\mathbf{y}$.

\subsubsection{Length Normalization and Orthogonal Transformation}

To ensure that all the training instances in the dictionary $\mathcal{D}$ contribute equally to the optimization objective in
(\ref{eq:bilingual-mapping}) and to preserve vector norms after projection, we have tried length normalization and orthogonal transformation for learning the bilingual mapping as in \cite{Xing2015,Artetxe2016,Smith2017}.

First, we normalize the source-language and target-language word embeddings to be unit vectors: $\mathbf{x}'=\frac{\mathbf{x}}{||\mathbf{x}||}$ for each source-language word embedding $\mathbf{x}$, and $\mathbf{y}'=
\frac{\mathbf{y}}{||\mathbf{y}||}$ for each target-language word embedding $\mathbf{y}$.

Next, we add an orthogonality constraint to (\ref{eq:bilingual-mapping}) such that $\mathbf{M}$ is an orthogonal matrix, i.e.,
$\mathbf{M}^\mathrm{T}\mathbf{M} = \mathbf{I}$ where $\mathbf{I}$ denotes the identity matrix:
\begin{equation} \label{eq:bilingual-mapping-orthogonal}
\mathbf{M}^{O} _{t\rightarrow s} = \arg\min_{\mathbf{M} \in \mathbb{R}^{d \times d}, \mathbf{M}^\mathrm{T}\mathbf{M} = \mathbf{I}}
\sum_{i=1}^{D}|| \mathbf{x}'_i - \mathbf{M}\mathbf{y}'_i||^2
\end{equation}
$\mathbf{M}^{O} _{t\rightarrow s}$ can be computed using singular-value decomposition (SVD).

\subsubsection{Semi-Supervised and Unsupervised Mappings}

The mapping learned in (\ref{eq:bilingual-mapping}) or (\ref{eq:bilingual-mapping-orthogonal}) requires a seed dictionary. To relax this requirement, \citet{Artetxe2017} proposed a self-learning procedure that can be combined with a dictionary-based mapping technique. Starting with a small seed dictionary, the procedure iteratively 1) learns a mapping using the current dictionary; and 2) computes a new dictionary using the learned mapping. 

\citet{Artetxe2018} proposed an unsupervised method to learn the bilingual mapping without using a seed dictionary. The method first uses a heuristic to build an initial dictionary that aligns the vocabularies of two languages, and then applies a robust self-learning procedure to iteratively improve the mapping. Another unsupervised method based on adversarial training was proposed in \citet{Conneau2018}.

We compare the performance of different mappings for cross-lingual RE model transfer in Section \ref{subsection:eval-mappings}.

\section{Neural Network RE Models} \label{section:neuralREmodels}

For any two entities in a sentence, an RE model determines whether these two entities have a relationship, and if yes, classifies the
relationship into one of the pre-defined relation types.  
We focus on neural network RE models since these models achieve the state-of-the-art performance for relation extraction. Most
importantly, neural network RE models use word embeddings as the input, which are amenable to cross-lingual model transfer via cross-lingual word embeddings. In this paper, we use English as the source language. 

Our neural network architecture has four layers. The first layer is the \emph{embedding layer} which maps input words in a sentence to
word embeddings. The second layer is a \emph{context layer} which transforms the word embeddings to context-aware vector representations
using a recurrent or convolutional neural network layer. The third layer is a \emph{summarization layer} which summarizes the vectors in
a sentence by grouping and pooling. The final layer is the \emph{output layer} which returns the classification label for the relation
type.

\subsection{Embedding Layer}

For an English sentence with $n$ words $\mathbf{s}=(w_1,w_2,...,w_n)$, the \emph{embedding layer} maps each word $w_t$ to a real-valued vector (word embedding) $\mathbf{x}_t\in \mathbb{R}^{d \times 1}$ using the English word embedding model (Section \ref{subsection:monolingual_embeddings}).
In addition, for each entity $m$ in the sentence, the
embedding layer maps its entity type to a real-valued vector (entity label embedding) $\mathbf{l}_m \in \mathbb{R}^{d_m \times 1}$ (initialized randomly). In our experiments we use $d=300$ and $d_m = 50$.

\subsection{Context Layer}

Given the word embeddings $\mathbf{x}_t$'s of the words in the sentence, the \emph{context layer} tries to build a
sentence-context-aware vector representation for each word. We consider two types of neural network layers that aim to achieve this.

\subsubsection{Bi-LSTM Context Layer}

The first type of context layer is based on Long Short-Term Memory (LSTM) type recurrent neural networks \cite{Hochreiter1997,Graves2015}. Recurrent neural networks (RNNs) are a class of neural networks that operate on sequential data such as sequences of words. LSTM networks are a type of RNNs that have been invented to better capture long-range dependencies in sequential
data.

We pass the word embeddings $\mathbf{x}_t$'s to a forward and a backward LSTM layer. A forward or backward LSTM layer
consists of a set of recurrently connected blocks known as \emph{memory blocks}. The memory block at the $t$-th word in the forward LSTM
layer contains a memory cell $\overrightarrow{\mathbf{c}}_t$ and three \emph{gates}: an input gate
$\overrightarrow{\mathbf{i}}_t$, a forget gate $\overrightarrow{\mathbf{f}}_t$ and an output gate $\overrightarrow{\mathbf{o}}_t$
($\overrightarrow{\cdot}$ indicates the forward direction), which are updated as follows:
\begin{eqnarray}
\overrightarrow{\mathbf{i}}_t & = & \sigma\big(\overrightarrow{W}_i \mathbf{x}_t + \overrightarrow{U}_i \overrightarrow{\mathbf{h}}_{t-1} + \overrightarrow{\mathbf{b}}_i\big) \nonumber \\
\overrightarrow{\mathbf{f}}_t & = & \sigma\big(\overrightarrow{W}_f \mathbf{x}_t + \overrightarrow{U}_f \overrightarrow{\mathbf{h}}_{t-1 } + \overrightarrow{\mathbf{b}}_f\big) \nonumber \\
\overrightarrow{\mathbf{o}}_t & = & \sigma\big(\overrightarrow{W}_o \mathbf{x}_t + \overrightarrow{U}_o \overrightarrow{\mathbf{h}}_{t-1 } + \overrightarrow{\mathbf{b}}_o\big) \nonumber \\
\overrightarrow{\mathbf{c}}_t & = & \overrightarrow{\mathbf{f}}_t \odot \overrightarrow{\mathbf{c}}_{t-1} + \nonumber \\
& & \overrightarrow{\mathbf{i}}_t \odot \tanh\big(\overrightarrow{W}_c \mathbf{x}_t + \overrightarrow{U}_c \overrightarrow{\mathbf{h}}_{t-1} + \overrightarrow{\mathbf{b}}_c\big) \nonumber \\
\overrightarrow{\mathbf{h}}_t & = & \overrightarrow{\mathbf{o}}_t \odot \tanh(\overrightarrow{\mathbf{c}}_t) 
\end{eqnarray}
where $\sigma$ is the element-wise
sigmoid function and $\odot$ is the element-wise multiplication.

The hidden state vector $\overrightarrow{\mathbf{h}}_t$ in the forward LSTM layer incorporates information from the left (past) tokens of $w_t$ in the
sentence. Similarly, we can compute the hidden state vector $\overleftarrow{\mathbf{h}}_t$ in the backward LSTM layer, which incorporates
information from the right (future) tokens of $w_t$ in the sentence. The concatenation of the two vectors $\mathbf{h}_t = [\overrightarrow{\mathbf{h}}_t,
\overleftarrow{\mathbf{h}}_t]$ is a good representation of the word $w_t$ with both left and right contextual information in the sentence.

\subsubsection{CNN Context Layer}

The second type of context layer is based on Convolutional Neural Networks (CNNs) \cite{Zeng2014,Santos2015}, which applies
\emph{convolution-like} operation on successive windows of size $k$ around each word in the sentence. Let $\mathbf{z}_t =
[\mathbf{x}_{t-(k-1)/2},...,\mathbf{x}_{t+(k-1)/2}]$ be the concatenation of  $k$ word embeddings around $w_t$. The convolutional layer
computes a hidden state vector
\begin{equation}
\mathbf{h}_t = \tanh(\mathbf{W} \mathbf{z}_t + \mathbf{b})
\end{equation}
for each word $w_t$, where $\mathbf{W}$ is a weight matrix and $\mathbf{b}$ is a bias vector, and $\tanh(\cdot)$ is the element-wise
hyperbolic tangent function.

\subsection{Summarization Layer}

After the context layer, the sentence $(w_1,w_2,...,w_n)$ is represented by $(\mathbf{h}_1,....,\mathbf{h}_n)$.  Suppose $m_1=(w_{b_1},..,w_{e_1})$ and
$m_2=(w_{b_2},..,w_{e_2})$ are two entities in the sentence where $m_1$ is on the left of $m_2$ (i.e., $e_1 < b_2$). As different
sentences and entities may have various lengths, the \emph{summarization layer} tries to build a \emph{fixed-length} vector that best
summarizes the representations of the sentence and the two entities for relation type classification.

We divide the hidden state vectors $\mathbf{h}_t$'s into 5 groups:
\begin{itemize}
\item $G_1=\{\mathbf{h}_{1},..,\mathbf{h}_{b_1-1}\}$ includes vectors that are left to the first entity $m_1$.

\item $G_2=\{\mathbf{h}_{b_1},..,\mathbf{h}_{e_1}\}$ includes vectors that are in the first entity $m_1$.

\item $G_3=\{\mathbf{h}_{e_1+1},..,\mathbf{h}_{b_2-1}\}$ includes vectors that are between the two entities.

\item $G_4=\{\mathbf{h}_{b_2},..,\mathbf{h}_{e_2}\}$ includes vectors that are in the second entity $m_2$.

\item $G_5=\{\mathbf{h}_{e_2+1},..,\mathbf{h}_{n}\}$ includes vectors that are right to the second entity $m_2$.
\end{itemize}

We perform element-wise max pooling among the vectors in each group:
\begin{equation}
\mathbf{h}_{G_i}(j)  = \max_{\mathbf{h} \in G_i} \mathbf{h}(j), 1\leq j\leq d_h, 1\leq i \leq 5
\end{equation}
where $d_h$ is the dimension of the hidden state vectors. Concatenating the $\mathbf{h}_{G_i}$'s we get a fixed-length vector
$\mathbf{h}_s=[\mathbf{h}_{G_1},...,\mathbf{h}_{G_5}]$.

\subsection{Output Layer}

The output layer receives inputs from the previous layers (the summarization vector $\mathbf{h}_s$, the entity label embeddings
$\mathbf{l}_{m_1}$ and $\mathbf{l}_{m_2}$ for the two entities under consideration) and returns a probability distribution over the
relation type labels:
\begin{equation}
\mathbf{p} = \textrm{softmax} \big(\mathbf{W}_s \mathbf{h}_s + \mathbf{W}_{m_1}\mathbf{l}_{m_1} + \mathbf{W}_{m_2}\mathbf{l}_{m_2} +
\mathbf{b}_o \big)
\end{equation}

\subsection{Cross-Lingual RE Model Transfer}

Given the word embeddings of a sequence of words in a target language $t$, $(\mathbf{y}_1,...,\mathbf{y}_n)$, we project them into the
English embedding space by applying the linear mapping $\mathbf{M}_{t\rightarrow s}$ learned in Section
\ref{subsection:cross-lingual-embedding-mapping}: $(\mathbf{M}_{t\rightarrow s}\mathbf{y}_1, \mathbf{M}_{t\rightarrow s}\mathbf{y}_2,...,\mathbf{M}_{t\rightarrow
s}\mathbf{y}_n)$. The neural network English RE model is then applied on the projected word embeddings and the entity label embeddings (which are language independent) to perform relationship classification.

Note that our models do not use language-specific resources such as dependency parsers or POS taggers because these resources might not
be readily available for a target language. Also our models do not use precise word position features since word positions in sentences
can vary a lot across languages.

\section{Experiments} \label{section:experiments}

In this section, we evaluate the performance of the proposed cross-lingual RE approach on both in-house dataset and the ACE (Automatic
Content Extraction) 2005 multilingual dataset \cite{Walker2006}.

\subsection{Datasets}

Our in-house dataset includes manually annotated RE data for 6 languages: English, German, Spanish, Italian, Japanese and Portuguese.
It defines 56 entity types (e.g., \emph{Person}, \emph{Organization}, \emph{Geo-Political Entity}, \emph{Location}, \emph{Facility}, \emph{Time},
\emph{Event{\_V}iolence}, etc.) and 53 relation types between the entities (e.g., \emph{AgentOf}, \emph{LocatedAt}, \emph{PartOf},
\emph{TimeOf}, \emph{AffectedBy}, etc.).

The ACE05 dataset includes manually annotated RE data for 3 languages: English, Arabic and Chinese. It defines 7 entity types
(\emph{Person}, \emph{Organization}, \emph{Geo-Political Entity}, \emph{Location}, \emph{Facility}, \emph{Weapon}, \emph{Vehicle}) and 6
relation types  between the entities (\emph{Agent-Artifact}, \emph{General-Affiliation}, \emph{ORG-Affiliation}, \emph{Part-Whole},
\emph{Personal-Social}, \emph{Physical}).

For both datasets, we create a class label ``O" to denote that the two entities under consideration do not have a relationship belonging to one of the relation types of interest.

\begin{figure*}
\includegraphics[scale=0.5]{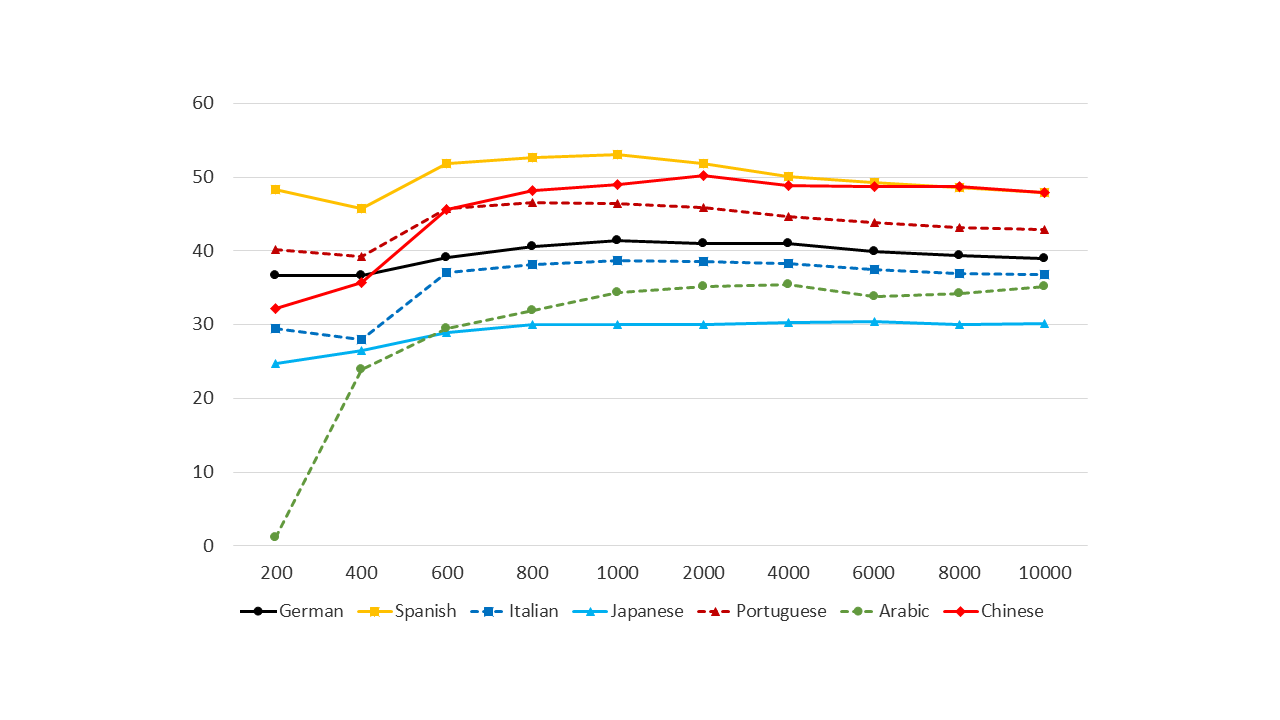} \caption{Cross-lingual RE performance ($F_1$ score) vs. dictionary size (number of bilingual word pairs for learning the mapping (\ref{eq:bilingual-mapping})) under the Bi-LSTM English RE model on the target-language development data.} \label{fig:dict-size}
\end{figure*}

\subsection{Source (English) RE Model Performance}

We build 3 neural network English RE models under the architecture described in Section \ref{section:neuralREmodels}:

\begin{itemize}

\item The first neural network RE model does not have a context layer and the word embeddings are directly passed to the summarization
layer. We call it \textbf{Pass-Through} for short.

\item The second neural network RE model has a Bi-LSTM context layer. We call it \textbf{Bi-LSTM} for short.

\item The third neural network model has a CNN context layer with a window size 3. We call it \textbf{CNN} for short.
\end{itemize}

\begin{table}
\footnotesize

\begin{center}
\begin{tabular}{|c|c|}

\hline \textbf{Model}  & $\mathbf{F_1}$ \\
\hline FCM (S) \cite{Gormley2015} & 55.06 \\
\hline Hybrid FCM (E) \cite{Gormley2015} & 58.26 \\
\hline BIDIRECT (S) \cite{Nguyen2016} & 57.73 \\
\hline VOTE-BW (E) \cite{Nguyen2016} & 60.60 \\
\hline Pass-Through (S) & 54.99 \\
\hline Bi-LSTM (S) & 58.92 \\
\hline CNN (S) & 57.91 \\
\hline
\end{tabular}

\end{center}
\caption{Comparison with the state-of-the-art RE models on the ACE05 English data (S: Single Model; E: Ensemble Model).} \label{table:comparison-ace05-en}
\end{table}

\begin{table}
\footnotesize
\begin{center}
\begin{tabular}{|c|c|c|c|}

\hline \textbf{In-House} & \textbf{Training} & \textbf{Dev} & \textbf{Test} \\
\hline English (Source)     & 1137 & 140 & 140 \\
\hline German (Target)      & 280 & 35 & 35  \\
\hline Spanish (Target)     & 451 & 55 & 55 \\
\hline Italian (Target)     & 322 & 40 & 40 \\
\hline Japanese (Target)    & 396 & 50 & 50 \\
\hline Portuguese (Target)  & 390 & 50 & 50 \\

\hline \textbf{ACE05} & \textbf{Training} & \textbf{Dev} & \textbf{Test} \\
\hline English (Source)  & 479 & 60 & 60 \\
\hline Arabic (Target)   & 323 & 40 & 40  \\
\hline Chinese (Target)  & 507 & 63 & 63 \\
\hline
\end{tabular}
\end{center}
\caption{Number of documents in the training/dev/test sets of the in-house and ACE05 datasets.}
\label{table:ace05-dataset-size}
\end{table}

First we compare our neural network English RE models with the state-of-the-art RE models on the ACE05 English data. The ACE05 English
data can be divided to 6 different domains: broadcast conversation (bc), broadcast news (bn), telephone conversation (cts), newswire
(nw), usenet (un) and webblogs (wl). We apply the same data split
in \cite{Plank2013,Gormley2015,Nguyen2016}, which uses news (the union
of bn and nw) as the training set, a half of bc as the development set and the remaining data as the test set.

We learn the model parameters using Adam \cite{Kingma2014}. We apply dropout \cite{Srivastava2014} to the hidden layers to reduce
overfitting. The development set is used for tuning the model hyperparameters and for early stopping.

In Table \ref{table:comparison-ace05-en} we compare our models with the best models in \cite{Gormley2015} and \cite{Nguyen2016}. Our
Bi-LSTM model outperforms the best model (single or ensemble) in \cite{Gormley2015} and the best single model in \cite{Nguyen2016},
without using any language-specific resources such as dependency parsers.

While the data split in the previous works was motivated by domain adaptation, the focus of this paper is on cross-lingual model transfer, and hence
we apply a random data split as follows. For the source language English and each target language, we randomly select $80\%$ of the data
as the training set, $10\%$ as the development set, and keep the remaining $10\%$ as the test set. The sizes of the sets are summarized
in Table \ref{table:ace05-dataset-size}.

We report the Precision, Recall and $F_1$ score of the 3 neural network English RE models in Table \ref{table:source-RE-model-performance}. Note that adding an additional context layer with either Bi-LSTM or CNN significantly improves the performance of our English RE model, compared with the simple Pass-Through model. Therefore, we will focus on the Bi-LSTM model and the CNN model in the subsequent experiments.

\begin{table}
\footnotesize
\begin{center}
\begin{tabular}{|c|p{4mm}p{4mm}p{4mm}|p{4mm}p{4mm}p{4mm}|}

\hline \textbf{Model} & \multicolumn {3}{|c|}{\textbf{In-House}} & \multicolumn {3}{|c|}{\textbf{ACE05}} \\
\cline{2-7} & \textbf{P} & \textbf{R} & $\mathbf{F_1}$ & \textbf{P} & \textbf{R} & $\mathbf{F_1}$   \\
\hline Pass-Through & 60.8 & 63.1 & 61.9   & 59.4 & 55.1 & 57.2  \\
\hline Bi-LSTM      & 66.5 & 67.7 & 67.1   & 65.1 & 65.8 & 65.5  \\
\hline CNN          & 63.4 & 68.8 & 66.0   & 61.7 & 67.1 & 64.3  \\
\hline
\end{tabular}
\end{center}
\caption{Performance of the supervised English RE models on the in-house and ACE05 English test data.}
\label{table:source-RE-model-performance}
\end{table}

\subsection{Cross-Lingual RE Performance}\label{CLRE}

We apply the English RE models to the 7 target languages across a variety of language families.

\subsubsection{Dictionary Size}

The bilingual dictionary includes the most frequent target-language words and their translations in English. To determine how many word pairs are needed to learn an effective bilingual word embedding mapping for cross-lingual RE, we first evaluate the performance ($F_1$ score) of our cross-lingual RE approach on the target-language development sets with an increasing dictionary size, as plotted in Figure \ref{fig:dict-size}.

We found that for most target languages, once the dictionary size reaches 1K, further increasing the dictionary size may not improve the transfer performance. Therefore, we select the dictionary size to be 1K.

\subsubsection{Comparison of Different Mappings} \label{subsection:eval-mappings}

\begin{table*}
\footnotesize
\begin{center}
\begin{tabular}{|c|c|c|c|c|c|c|c|c|}

\hline \textbf{Mapping} & \textbf{German} & \textbf{Spanish} & \textbf{Italian} & \textbf{Japanese} & \textbf{Portuguese} & \textbf{Arabic} & \textbf{Chinese} & \textbf{Average} \\
\hline Regular-1K    & 41.4 & 53.0 & 38.7 & 30.0 & 46.4 & 34.4 & 49.0 & 41.8 \\
\hline Orthogonal-1K    & 41.0 & 49.2 & 35.7 & 27.1 & 44.0 & 32.1 & 47.2 & 39.5\\
\hline Semi-Supervised-1K   & 38.1 & 46.7 & 31.2 & 25.3 & 43.1 & 37.3 & 47.1 & 38.4 \\
\hline Unsupervised   & 37.8 & 45.5 & 28.9 & 21.2 & 40.8 & 37.5 & 49.1 & 37.3 \\
\hline
\end{tabular}
\end{center}
\caption{Comparison of the performance ($F_1$ score) using different mappings on the target-language development data under the Bi-LSTM model.} \label{table:compare-mappings}
\end{table*}

We compare the performance of cross-lingual RE model transfer under the following bilingual word embedding mappings:
\begin{itemize}
    
\item \textbf{Regular-1K}: the regular mapping learned in (\ref{eq:bilingual-mapping}) using 1K word pairs;

\item \textbf{Orthogonal-1K}: the orthogonal mapping with length normalization learned in (\ref{eq:bilingual-mapping-orthogonal}) using 1K word pairs (in this case we train the English RE models with the normalized English word embeddings);

\item \textbf{Semi-Supervised-1K}: the mapping learned with 1K word pairs and improved by the self-learning method in \cite{Artetxe2017};

\item \textbf{Unsupervised}: the mapping learned by the unsupervised method in \cite{Artetxe2018}.
    
\end{itemize}

The results are summarized in Table \ref{table:compare-mappings}. The regular mapping outperforms the orthogonal mapping consistently across the target languages. While the orthogonal mapping was shown to work better than the regular mapping for the word translation task \cite{Xing2015,Artetxe2016,Smith2017}, our cross-lingual RE approach directly maps target-language word embeddings to the English embedding space without conducting word translations. Moreover, the orthogonal mapping requires length normalization, but we observed that length normalization adversely affects the performance of the English RE models (about 2.0 $F_1$ points drop).

We apply the vecmap toolkit\footnote{https://github.com/artetxem/vecmap} to obtain the semi-supervised and unsupervised mappings. The unsupervised mapping has the lowest average accuracy over the target languages, but it does not require a seed dictionary. Among all the mappings, the regular mapping achieves the best average accuracy over the target languages using a dictionary with only 1K word pairs, and hence we adopt it for the cross-lingual RE task.

\begin{table*}
\footnotesize\centering
\begin{center}
\begin{tabular}{|c|p{4mm}p{4mm}p{4mm}|p{4mm}p{4mm}p{4mm}|p{4mm}p{4mm}p{4mm}|p{4mm}p{4mm}p{4mm}|p{4mm}p{4mm}p{4mm}|}

\hline \textbf{Model}  & \multicolumn {3}{|c|}{\textbf{German}} & \multicolumn {3}{|c|}{\textbf{Spanish}} & \multicolumn
{3}{|c|}{\textbf{Italian}} & \multicolumn{3}{|c|}{\textbf{Japanese}} & \multicolumn {3}{|c|}{\textbf{Portuguese}} \\
\cline{2-16} & \textbf{P} & \textbf{R} & $\mathbf{F_1}$ & \textbf{P} & \textbf{R} & $\mathbf{F_1}$ & \textbf{P} & \textbf{R} & $\mathbf{F_1}$ & \textbf{P} & \textbf{R} & $\mathbf{F_1}$ & \textbf{P} & \textbf{R} & $\mathbf{F_1}$\\
\hline Bi-LSTM       & 39.6 & 48.9 & 43.8  & 54.5 & 47.6 & 50.8   & 41.8 & 34.2 & 37.6  & 33.9 & 25.1 & 28.9  & 52.9 & 44.5 & 48.4 \\
\hline CNN           & 32.5 & 50.5 & 39.5  & 49.3 & 48.3 & 48.8   & 36.6 & 34.9 & 35.7  & 27.3 & 31.5 & 29.3  & 49.0 & 44.0 & 46.3 \\
\hline Ensemble      & 39.6 & 50.5 & 44.4  & 56.9 & 49.1 & 52.7   & 42.6 & 35.3 & 38.6  & 35.3 & 26.4 & 30.2  & 54.9 & 45.2 & 49.6 \\
\hline \emph{Supervised}    & \emph{59.3} & \emph{56.4} & \emph{57.8}  & \emph{68.4} & \emph{65.4} & \emph{66.8}   & \emph{51.4} & \emph{48.3} & \emph{49.8}  & \emph{52.7} & \emph{52.0} & \emph{52.4}  & \emph{64.0} & \emph{61.3} & \emph{62.6} \\

 \hline
\end{tabular}
\end{center}
\caption{Performance of the cross-lingual RE approach on the in-house target-language test data.}
\label{table:transfer-evaluation-inhouse}
\end{table*}

\begin{table}
\footnotesize
\begin{center}

\begin{tabular}{|c|p{4mm}p{4mm}p{4mm}|p{4mm}p{4mm}p{4mm}|}

\hline \textbf{Model}  & \multicolumn {3}{|c|}{\textbf{Arabic}} & \multicolumn {3}{|c|}{\textbf{Chinese}} \\
\cline{2-7} & \textbf{P} & \textbf{R} & $\mathbf{F_1}$ & \textbf{P} & \textbf{R} & $\mathbf{F_1}$ \\
\hline         Bi-LSTM  & 30.3 & 45.7 & 36.4  &  61.7 & 37.8 & 46.8 \\
\hline             CNN  & 24.0 & 39.7 & 29.9  &  56.4 & 33.8 & 42.3 \\
\hline        Ensemble  & 27.5 & 48.7 & 35.2  &  61.0 & 40.4 & 48.6 \\
\hline      \emph{Supervised}  & \emph{70.0} & \emph{69.1} & \emph{69.5}  &  \emph{66.9} & \emph{69.4} & \emph{68.1} \\
\hline
\end{tabular}

\end{center}
\caption{Performance of the cross-lingual RE approach on the ACE05 target-language test data.} \label{table:transfer-evaluation-ace05}
\end{table}

\subsubsection{Performance on Test Data}

The cross-lingual RE model transfer results for the in-house test data are summarized in Table \ref{table:transfer-evaluation-inhouse} and the
results for the ACE05 test data are summarized in Table \ref{table:transfer-evaluation-ace05}, using the regular mapping learned with a bilingual dictionary of size 1K. In the tables, we also provide the performance of the supervised RE model (Bi-LSTM) for each target language, which is trained with a few hundred thousand tokens of manually annotated RE data in the target-language, and may serve as an upper bound for the cross-lingual model transfer performance.

Among the 2 neural network models, the Bi-LSTM model achieves a better cross-lingual RE performance than the CNN model for 6 out of the 7 target languages. In terms of \emph{absolute} performance, the Bi-LSTM model achieves over $40.0$ $F_1$ scores for German, Spanish, Portuguese and Chinese. In terms of \emph{relative} performance, it reaches over $75\%$ of the accuracy of the supervised target-language RE model for German, Spanish, Italian and Portuguese. While Japanese and Arabic appear to be more difficult to transfer, it still achieves $55\%$ and $52\%$ of the accuracy of the supervised Japanese and Arabic RE model, respectively, without using any manually annotated RE data in Japanese/Arabic.

We apply model ensemble to further improve the accuracy of the Bi-LSTM model. We train 5 Bi-LSTM English RE models initiated with
different random seeds, apply the 5 models on the target languages, and combine the outputs by selecting the relation type labels with the
highest probabilities among the 5 models. This \textbf{Ensemble} approach improves the single model by 0.6-1.9 $F_1$ points, except for Arabic.

\subsubsection{Discussion}

Since our approach projects the target-language word embeddings to the source-language embedding space preserving the word order, it is expected to work better for a target language that has more similar word order as the source language. This has been verified by our experiments. The source language, English, belongs to the SVO (Subject, Verb, Object) language family where in a sentence the subject comes first, the verb second, and the object third. Spanish, Italian, Portuguese, German (in conventional typology) and Chinese also belong to the SVO language family, and our approach achieves over $70\%$ relative accuracy for these languages. On the other hand, Japanese belongs to the SOV (Subject, Object, Verb) language family and Arabic belongs to the VSO (Verb, Subject, Object) language family, and our approach achieves lower relative accuracy for these two languages.

\section{Related Work}

There are a few weakly supervised cross-lingual RE approaches.  \citet{Kim2010} and \citet{Kim2012} project annotated English RE data to Korean to create weakly labeled training data via aligned parallel corpora. \citet{Faruqui2015} translates a target-language sentence into English, performs RE in English, and then projects the relation phrases back to the target-language sentence. \citet{Zou2018} proposes an adversarial feature adaptation approach for cross-lingual relation classification, which uses a machine translation system to translate source-language sentences into target-language sentences. 
Unlike the existing approaches, our approach does
not require aligned parallel corpora or machine translation systems. There are also several multilingual RE approaches, e.g.,
\citet{Verga2016,Min2017,Lin2017}, where the focus is to improve monolingual RE by jointly modeling texts in multiple languages.

Many cross-lingual word embedding models have been developed recently \cite{Upadhyay2016,Ruder2017}. An important application of cross-lingual word embeddings is to enable cross-lingual model transfer. In this paper, we apply the bilingual word embedding mapping  technique in \cite{Mikolov2013b} to cross-lingual RE model transfer. Similar approaches have been applied to
other NLP tasks such as dependency parsing \cite{Guo2015}, POS tagging \cite{Gouws2015b} and named entity recognition \cite{Ni2017,Xie2018}.

\section{Conclusion}

In this paper, we developed a simple yet effective neural cross-lingual RE model transfer approach, which has very low resource requirements (a small bilingual dictionary with 1K word pairs) and can be easily extended to a new language. Extensive experiments for 7 target languages across a variety of language families on both in-house and open datasets show that the proposed approach achieves very good performance (up to $79\%$ of the accuracy of the supervised target-language RE model), which provides a strong baseline for building cross-lingual RE models with minimal resources.

\section*{Acknowledgments}

We thank Mo Yu for sharing their ACE05 English data split and the anonymous reviewers for their valuable comments.

\bibliography{cross-lingual-re}
\bibliographystyle{acl_natbib}

\end{document}